\documentclass{article}

     \usepackage[final]{neurips_2021}

\usepackage{xcolor}         % colors
\usepackage[utf8]{inputenc} % allow utf-8 input
\usepackage[T1]{fontenc}    % use 8-bit T1 fonts
\definecolor{mydarkblue}{rgb}{0,0.08,0.45}
\usepackage[colorlinks,citecolor=mydarkblue,urlcolor=mydarkblue,linkcolor=mydarkblue]{hyperref}
\usepackage{url}            % simple URL typesetting
\usepackage{booktabs}       % professional-quality tables
\usepackage{amsfonts}       % blackboard math symbols
\usepackage{nicefrac}       % compact symbols for 1/2, etc.
\usepackage{microtype}      % microtypography
\usepackage{soul}
\usepackage{enumitem}
\usepackage{graphicx}
\usepackage{subcaption}
\usepackage{bbm}

\usepackage[compact]{titlesec}
\titlespacing{\section}{0pt}{2ex}{1ex}
\titlespacing{\subsection}{0pt}{1ex}{0ex}
\titlespacing{\subsubsection}{0pt}{0.5ex}{0ex}

\title{
Reliable Graph Neural Networks for Drug Discovery Under Distributional Shift}

\author{%
  Kehang Han \thanks{Google AI Resident} \thanks{Corresponding author} \\
  Google Research\\
  \texttt{kehanghan@google.com} \\
  \And
  Balaji Lakshminarayanan \\
  Google Research\\
  \texttt{balajiln@google.com} \\
  \And
  Jeremiah Liu \footnotemark[2]\\
  Google Research\\
  \texttt{jereliu@google.com} \\
}

\begin{document}

\maketitle

\begin{abstract}
  The concern of overconfident mispredictions under distributional shift demands extensive reliability research on Graph Neural Networks used in critical tasks in drug discovery. Here we first introduce \textbf{CardioTox}, a real-world benchmark on drug cardiotoxicity to facilitate such efforts. Our exploratory study shows overconfident mispredictions are often distant from training data. That leads us to develop distance-aware GNNs: \textbf{GNN-SNGP}. Through evaluation on CardioTox and three established benchmarks, we demonstrate GNN-SNGP's effectiveness in increasing distance-awareness, reducing overconfident mispredictions and making better calibrated predictions without sacrificing accuracy performance. Our ablation study further reveals the representation learned by GNN-SNGP improves distance-preservation over its base architecture and is one major factor for improvements.
\end{abstract}

\section{Introduction}

In recent years, Graph Neural Networks (GNNs) have illustrated remarkable performance in scientific applications such as physics \citep{Battaglia2016-qj,Sanchez-Gonzalez2018-fs}, biomedical science \citep{Fout_undated-yf}, and computational chemistry \citep{Duvenaud2015-zr, Kearnes2016-sk}. One important example is the early-phase drug discovery, where GNNs have shown promise in critical tasks such as \emph{hit-finding} and \emph{liability screening} (i.e., predicting the \emph{binding affinity} and the \emph{toxicity} of candidate drug molecules, respectively) \citep{McCloskey2020-es, Siramshetty2020-yy}. 

However, a key reliability concern that hinders the adoption of GNN in real practice is the \emph{overconfident mispredictions}. For example, in liability screening, an Overconfident False Negative (OFN hereafter) prediction leads the GNN model to mark a toxic molecule as safe, causing severe consequences by leaking it to the next stage of drug development. Such concern is further exacerbated by a second key characteristic of the drug discovery tasks: \emph{data distributional shift}; 
drug discovery tasks often explicitly evaluate novel molecules by moving into regions in the feature space that are not previously represented in training data. As a result, the testing molecules are characteristically distinct from the training data, and can carry novel toxic signatures that were previously unseen by the model. A model with just high in-distribution accuracy is not sufficient under those situations; quantifying model reliability and designing techniques to improve robustness against overconfidence under distributional shift become especially relevant.

Currently, there lacks a drug discovery benchmark that targets realistic concerns about model reliability under distributional shifts. Thus, we introduce \textbf{CardioTox}, a data benchmark based on a real-world drug discovery problem and is compiled from 9K+ drug-like molecules from ChEMBL, NCATS and FDA validation databases \citep{Siramshetty2020-yy}. To evaluate model reliability, we generate additional molecular annotations and propose novel metrics to measure the models against the real-world standards around the responsible application of GNN. This is our first contribution.

Using CardioTox, our second contribution is an exploratory study on the root causes behind the overconfident mispredictions under distributional shift, and principled modeling approaches to mitigate it. In particular, we observe that many  overconfidently mispredicted molecules are structurally distinct from training data (i.e., they are "far" from training data based on molecule-fingerprint graph distance \citep{Rogers2010-yh}, see Section \ref{new_benchmark}). This failure mode suggests that improving the \textit{distance-awareness} of a GNN model would be an effective solution: a test molecule that is far away from the decision boundary (e.g., a toxic but novel molecule, due to its novelty, may possess few prior toxic signatures defining the decision boundary) should still get low confidence if it's distant from training data.  To this end, the recently proposed Spectral-normalized Neural Gaussian Processes (SNGP) \citep{liu2020simple} demonstrates a concrete approach to achieve this goal. Specifically, it imposes a distance-preserving regularization (i.e., spectral normalization) to the feature extractor, and replaces the dense output layer with a distance-aware classifier (i.e., random-feature Gaussian process). SNGP has shown promising robustness improvements in vision and language problems. To our best knowledge, this is the first study to bring the distance-aware design principle to GNN models to to improve reliability  performance on the molecule graph.

To summarize, our contributions are the following (Appendix  \ref{related_work} summarizes the related work):

\begin{itemize}[leftmargin=1em,itemsep=0.5em]

\item \textbf{Data}: We introduce \textbf{CardioTox}, a real-world drug discovery dataset with multiple test sets (IID and distribution-shifted ones), to the robustness community to facilitate reliability research of graph models.
The distributional shift challenge reflected in CardioTox is realistically faced by the field: test domain often comes from a data source that's different from the training source and has considerable amount of novel molecule graph structures (see also  Figure \ref{fig:data-split} in Appendix~\ref{cardiotox_split}). We further design distance-based data splits for CardioTox to quantitatively measure model's distance-awareness.

\item \textbf{Model}: We develop an end-to-end trainable \textbf{GNN-SNGP} architecture as well as its ablated version GNN-GP. Empirically, this method outperforms its base architecture in not only accuracy (e.g., AUROC) but also robustness (e.g., ECE) especially in reducing overconfident mispredictions. Measured by CardioTox's distance-based data splits, GNN-SNGP shows higher distance-awareness under data shifts than the  baselines, which explains its ability in overconfident misprediction reduction. Our implementation together with CardioTox dataset will be open sourced via Github.\footnote{\url{https://github.com/google/uncertainty-baselines/tree/main/baselines/drug_cardiotoxicity}} 

\item \textbf{Ablation Study}: We carry out extensive ablation studies which confirm that the robustness improvements come from both distance-aware classifier (GP-layer) and the distance-preserving latent representations. We also investigate the generalizability of this approach by testing on other graph modeling domain: molHIV, BBBP and BACE and obtained consistent results.

\end{itemize}

\vspace{-1em}
\section{Methods}
\label{methods}

\vspace{-0.5em}
\paragraph{GNN baseline}
We use a vanilla Message Passing Neural Network (MPNN) \citep{Gilmer2017-yn} as the GNN baseline in this study. Specifically, the message function is modeled by a dense layer $M(h_v, h_w, e_{vw}) = W_1 [h_v || h_w || e_{vw}]$, where $h_v, h_w$ are node features for node $v, w$ respectively, $e_{vw}$ is the edge feature vector between nodes $v, w$, and $W_1$ is the weight matrix. After getting aggregated message $m_v$ for each node, the hidden node feature $h_v$ is updated via a Gated Recurrent Unit $U(h_v, m_v) = \mathsf{GRU}(h_v, m_v)$. Finally, we read out graph level representation by $R = \sum_{v\in V} \sigma (W_2 [h_v || h_v^0]) \odot (W_3 h_v)$, where $W_2$, $W_3$ are two weight matrices and $\sigma$ is the sigmoid function. The graph level representation is fed into a final dense layer to generate logits.
\vspace{-0.5em}
\paragraph{GNN-GP: improving distance-awareness of classifier}
Following \citet{liu2020simple} who proposed Gaussian process layer (GP-layer hereafter) in vision and language models, we introduce it to the graph domain and developed GNN-GP model to increase distance-awareness. Figure \ref{fig:gnn-gp-architecture1} (Appendix \ref{architecture}) shows the high-level architectural changes.

\begin{figure}[ht]
    \centering
    \includegraphics[width=0.75\linewidth]{"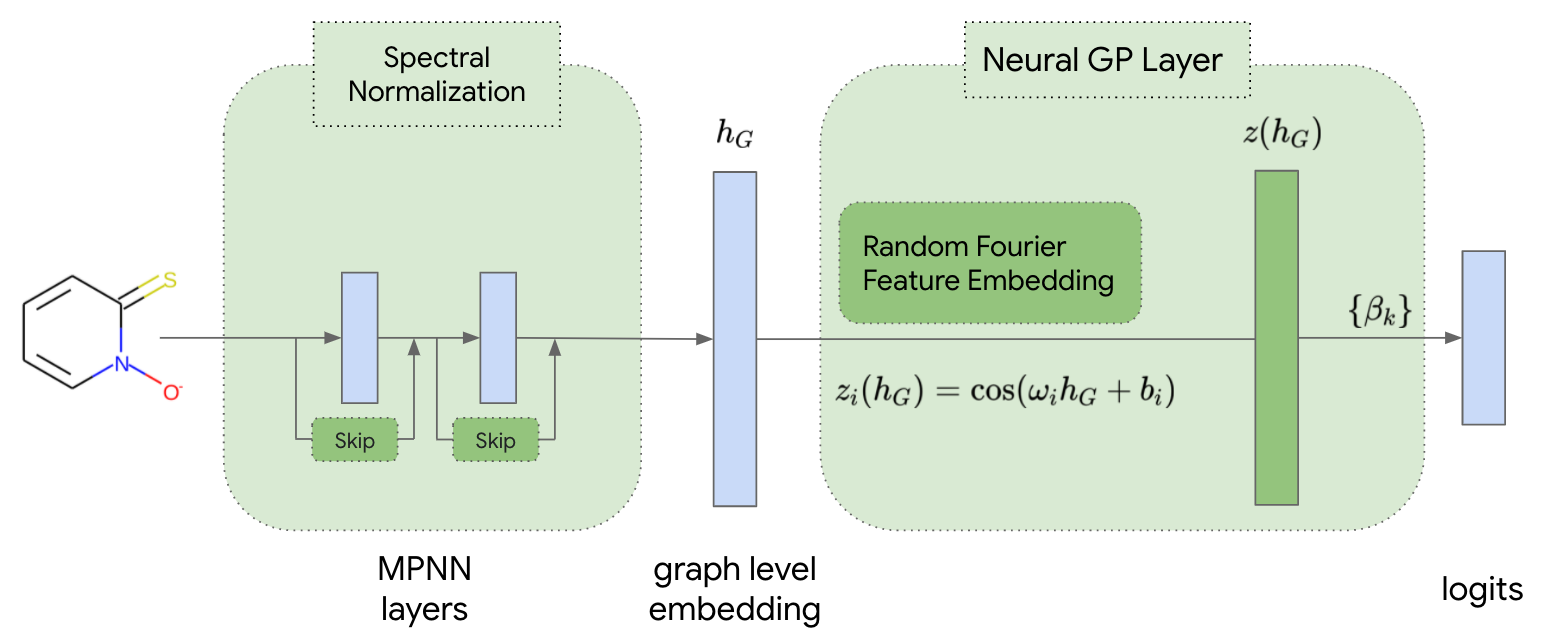"}
    \caption{Improving GNN architecture via distance-preserving feature extractor (skip connection and spectral normalization) and distance-aware classifier (neural Gaussian Processes layer).}
    \label{fig:gnn-gp-architecture}
\end{figure}

\vspace{-0.5em}
As Figure \ref{fig:gnn-gp-architecture} shows in detail, Gaussian processes are made end-to-end trainable with GNN by approximating GP kernel function via random Fourier feature generation \citep{Rahimi_undated-sq}. Specifically, coefficients $\omega_i, b_i$ are randomly sampled at initialization and kept fixed during training. The distribution that $\omega_i, b_i$ are sampled from depends on what Gaussian processes kernel we’d like to approximate. Take the Gaussian kernel as an example, we have $\omega_i \sim N(0, 1); b_i \sim U(0, 2\pi)$. 
During inference time, each sample would get logit predictions as well logit variances, both of which are utilized to compute predictive probabilities via mean-field approximation \citep{lu2020uncertainty}.

\vspace{-0.5em}
\paragraph{GNN-SNGP: incorporating distance-preserving feature extraction}
Due to feature collapse in feature extraction \citep{liu2020simple,Van_Amersfoort2021-qt}, neural representation may not faithfully preserve distance in the input manifold.
\citet{liu2020simple} propose to preserve input distance in feature extraction by applying Spectral Normalization (SN) \citep{Gouk2018-ia, Miyato2018-bg}  to the residual networks. As a result, combining SN and GP would increase the model's overall distance-awareness, helping OOD detection and other robustness metrics.

Since our vanilla MPNN model (GNN baseline) does not have residual connections, we create GNN-SNGP through two following changes (Figure \ref{fig:gnn-gp-architecture}). First, we model the message function via a dense layer with SN: $M(h_v, h_w, e_{vw}) = W_1^{SN} [h_v || h_w || e_{vw}] $
where $W_1^{SN}$ is the dense layer weight matrix whose spectral norm is regularized. Second, we add residual connection to the message passing layer through the node update function: $U(h_v, m_v) = \mathsf{GRU}(h_v, m_v) + h_v$.

\section{CardioTox: Drug cardiotoxicity under distributional shift}
\label{new_benchmark}

% \jzl{This section can become a main section (rather than subsubsection). Also can consider rubbing "under distributional shift" into the dataset name (to improve relevance to the venue, also since its an unique aspect of this dataset).}

\paragraph{Dataset information} In early drug discovery stages, two types of models are widely used: hit-finding model and anti-target (i.e., liability) model. The former scores a molecule based on its potential of making a drug (e.g., whether a molecule could bind tightly to the disease-causing protein target), while the latter aims to filter out those molecules with potentially high liability (e.g., whether a molecule could be toxic). One such liability is cardiotoxicity, which occurs when a molecule inhibits hERG, a protein target that is related to heart-rhythm control \citep{Smith1996-df, Vandenberg2012-gr}. Due to previous failing incidents, FDA has required new drugs to pass drug cardiotoxicity examination \citep{Center_for_Drug_Evaluation_undated-ql}.

Thus in this work, we introduce CardioTox, a benchmark based on a real-world drug discovery problem and is compiled from 9K+ drug-like molecules from ChEMBL and NCATS  databases \citep{Siramshetty2020-yy}. To evaluate GNN model reliability, we add graph structural information (e.g., node features) and set up three test sets that reflect distributional shifts: Test-IID is a set that’s sampled from the same distribution as the Train set. Test-OOD1 and Test-OOD2 are molecule sets coming from NCATS and FDA respectively: 84\% of Test-OOD1 and 82\% of Test-OOD2 are novel molecules that are distant from Train set (Figure \ref{fig:data-split}). We further generate additional molecular annotations: \textit{close} sample or \textit{far} sample based on Tanimoto fingerprint distance \citep{Bajusz2015-oz} to the train set (Appendix \ref{cardiotox_split})
. This setup allows us to quantitatively assess models' distance-awareness.

\textbf{Evaluation metrics}
We evaluate popular metrics for uncertainty such as  Expected Calibration Error (ECE), Brier score (Brier), Negative Log Likelihood (NLL) (cf. \citep{Ovadia2019-by} for an overview). To assess the models' abilities in capturing graph distance and mitigating over-confidence, we also introduce two new metrics: \textit{Percentage of Over-confident False Negatives} (OFNs \%): $\frac{\sum_i \mathbbm{1}(\hat{y}_i<0.1 \hspace{0.1cm} \mathsf{and} \hspace{0.1cm} y_i^\mathsf{true}=1)}{\sum_i \mathbbm{1}(\hat{y}_i<0.1)}$ where $\hat{y}_i, y_i^\mathsf{true}$ are predictive probability and label for the positive class (i.e., toxic class in CardioTox) for the $i$-th test sample, and \textit{Distance-Awareness AUC} (DA-AUC) that evaluates if the model assigns higher confidence to test inputs in \textit{close set} than those in \textit{far set} (find details Appendix \ref{metric_def}). Note that OFNs \% can be "cheated" by making all predictions under-confident, therefore an ideal uncertainty model should perform well \textit{both} in calibration metrics and in OFNs \%.

\textbf{Accuracy and robustness performance}
Table \ref{drug_cardiotoxicity_results} shows that GNN-GP outperforms GNN baseline in both AUROC and robustness metrics for the CardioTox task. This is the case for both the in-distribution (Test-IID) and the shifted test sets (Test-OOD1 and Test-OOD2). GNN-SNGP shows additional gains in robustness. If resource allows, Deep Ensemble \citep{lakshminarayanan2016simple} of GNN-SNGP further boosts performance, eliminating all OFNs in Test-OOD2. It is worth noting that our GNN-SNGP ensemble has outperformed previous state-of-the-art neural models \citep{Siramshetty2020-yy} in AUROC performance.

\begin{table}[h]
  \caption{Accuracy (AUROC), robustness (ECE, Brier, NLL) and overconfidence (OFNs) performance for Drug cardiotoxicity benchmark. Results are averaged over 10 seeds.}
  \label{drug_cardiotoxicity_results}
  \centering
  \resizebox{0.85\textwidth}{!}{  
  \begin{tabular}{llllllll}
    \toprule
    Test-IID & AUROC ($\uparrow$) & ECE ($\downarrow$) & Brier ($\downarrow$) & NLL ($\downarrow$) & OFNs\% ($\downarrow$) & DA-AUC ($\uparrow$) \\
    \midrule
    GNN baseline & 0.919$\pm$0.003  & 0.037$\pm$0.003 & 0.194$\pm$0.007 & 0.352$\pm$0.018 & 4.05$\pm$0.26  &  0.500$\pm$0.004 \\
    GNN-GP     & \textbf{0.937$\pm$0.001} & 0.036$\pm$0.002 & \textbf{0.176$\pm$0.008} & \textbf{0.298$\pm$0.015} & \textbf{3.51$\pm$0.17} & \textbf{0.523$\pm$0.004}     \\
    GNN-SNGP     & 0.932$\pm$0.001 & \textbf{0.028$\pm$0.001} & \textbf{0.179$\pm$0.005} & \textbf{0.295$\pm$0.005} & \textbf{3.56$\pm$0.21} & 0.517$\pm$0.004 \\
    \midrule
    GNN-SNGP Ensemble    & \textbf{0.942$\pm$0.000} & \textbf{0.013$\pm$0.001} & \textbf{0.162$\pm$0.000} & \textbf{0.264$\pm$0.001} & \textbf{2.54$\pm$0.07} & \textbf{0.525$\pm$0.002}  \\
    \bottomrule
    \toprule
    Test-OOD1 &  AUROC ($\uparrow$) & ECE ($\downarrow$) & Brier ($\downarrow$) & NLL ($\downarrow$) & OFNs\% ($\downarrow$) & DA-AUC ($\uparrow$) \\
    \midrule
    GNN baseline & 0.786$\pm$0.004  & 0.102$\pm$0.012 & 0.343$\pm$0.071  & 0.578$\pm$0.125 & 1.68$\pm$0.08 &  0.604$\pm$0.006  \\
    GNN-GP     & 0.823$\pm$0.003 & 0.090$\pm$0.010 & 0.327$\pm$0.061 & 0.546$\pm$0.107 & 1.41$\pm$0.12 &  \textbf{0.632$\pm$0.002}  \\
    GNN-SNGP     & \textbf{0.836$\pm$0.003} & \textbf{0.074$\pm$0.008} & \textbf{0.316$\pm$0.047} & \textbf{0.503$\pm$0.072} & \textbf{1.31$\pm$0.09} & \textbf{0.635$\pm$0.007} \\
    \midrule
    GNN-SNGP Ensemble    & \textbf{0.850$\pm$0.002} & \textbf{0.039$\pm$0.002} & \textbf{0.277$\pm$0.005} & \textbf{0.428$\pm$0.006} & \textbf{1.22$\pm$0.08} & \textbf{0.643$\pm$0.003}  \\
    \bottomrule
    \toprule
    Test-OOD2 &  AUROC ($\uparrow$) & ECE ($\downarrow$) & Brier ($\downarrow$) & NLL ($\downarrow$) & OFNs\% ($\downarrow$) & DA-AUC ($\uparrow$) \\
    \midrule
    GNN baseline & 0.831$\pm$0.007  & 0.082$\pm$0.012 & 0.284$\pm$0.060 & 0.492$\pm$0.113 & 1.73$\pm$0.20 &  0.630$\pm$0.008 \\
    GNN-GP     & 0.873$\pm$0.006 & 0.074$\pm$0.007 & 0.261$\pm$0.047 & 0.442$\pm$0.086 & \textbf{0.98$\pm$0.19} &  0.656$\pm$0.011   \\
    GNN-SNGP     & \textbf{0.885$\pm$0.007} & \textbf{0.044$\pm$0.006} & \textbf{0.238$\pm$0.040} & \textbf{0.389$\pm$0.068} & \textbf{1.02$\pm$0.11} & \textbf{0.678$\pm$0.008}  \\
    \midrule
    GNN-SNGP Ensemble    & \textbf{0.896$\pm$0.002} & \textbf{0.021$\pm$0.002} & \textbf{0.210$\pm$0.003} & \textbf{0.333$\pm$0.005} & \textbf{1.06$\pm$0.11} & \textbf{0.682$\pm$0.003}  \\
    \bottomrule
  \end{tabular}
  }
\end{table}

\vspace{0.2em}
\textbf{Why does distance-awareness help reduce overconfident mispredictions?}
We observe a significant portion of OFNs are distant from the train set (i.e., 60\% of them have Tanimoto distance $>0.65$ \citep{Bajusz2015-oz}, also see Figure \ref{fig:ofn_distance_cdf}).
This could happen when a GNN model lacks of distance-awareness: a toxic but novel molecule can be far away from the wrong side of model's decision boundary, due to lacking known toxic signatures that is present in train set. Without being aware of the distance to train set, the GNN baseline model tends to base its prediction on distance to the decision boundary and give high confidence for such cases.

To this end, GNN-GP leverages GP's distance-awareness and is able to naturally incorporate distance into predictive uncertainty. As shown in Table \ref{drug_cardiotoxicity_results}, the SN version GNN-SNGP achieves highest distance-awareness (DA-AUC) in the two data-source-shifted test sets, correlating well with OFNs reduction. Figure \ref{fig:ofn_distance_cdf} shows a decreasing trend (from GNN baseline to GNN-GP to GNN-SNGP) of the percentage of distant samples among OFNs. Overall, using GP is able to improve uncertainty estimate for over 80\% of the baseline OFNs while also improving calibration (Figure \ref{fig:uncertainty_increase_ratio}).

\vspace{-0.1em}
\textbf{Additional ablations to assess relative contributions of SN and GP} %Performance contribution analysis}
In order to understand relative performance contributions of the two modeling components (i.e., distance-preserving feature extractor vs distance-aware classifier), we carry out an extensive ablation study in Appendix \ref{ablation_results_section}. We take the latent representations learned by GNN baseline, GNN-GP and GNN-SNGP (increasing distance-preservation) and feed them to classifiers with increasing distance-awareness: Dense layer, GP-layer, exact Gaussian processes classifier (GPC hereafter). Table \ref{ablation_results} suggests there's synergy between distance-preservation of neural representation and distance-awareness of the classifier. With the least distance-aware classifier (i.e., Dense layer), AUROC drops when increasing distance-preservation in neural representation. With the most distance-aware classifier (i.e., exact GPC), increasing neural representation's distance-preservation benefits accuracy, robustness as well as overconfidence reduction. We find a GNN model achieves best performances when both are present in the architecture (GNN-SNGP embeddings with GPC). Interesting, naively increasing distance-preservation along is not sufficient in guaranteeing good  generalization; we observe that the pre-defined representation (i.e., the molecule fingerprint FP) gives relatively low AUROC 
%(compared with neural representation)
under data shifts (e.g., on Test-OOD2) despite being perfectly distance-preserving. This is related to the trade-off in representation learning between dimension reduction and information preservation. Appendix \ref{ablation_results_section} discusses in further detail.

\vspace{-0.1em}
\textbf{Additional results on existing benchmarks}
Consistent with the results obtained in CardioTox, GNN-GP also outperforms GNN baseline on three established graph classification benchmarks \citep{Wu2017-yv}: molHIV, BBBP and BACE (see Appendix \ref{established_benchmarks}). We can make a few observations on the results in Table \ref{benchmarks_results}. First, GNN-GP consistently achieves higher AUROC, improves calibration as well as reduces overconfident mispredictions across all the benchmarks than the baseline. Second, with limited accuracy performance drop, the spectral normalized version GNN-SNGP can further improve model robustness compared with GNN-GP.

\section{Conclusion}
\label{conclusion}
In this study, we introduce GNN-GP and GNN-SNGP together with a new benchmark CardioTox from drug discovery setting. Through evaluation on four datasets, we demonstrate their effectiveness in reducing overconfident mispredictions and making better calibrated GNN models without sacrificing accuracy performance. The improved robustness appears to come from the boost in distance-awareness. We further discover that the embedding space induced by SN and GP addition improves distance-preservation over its base architecture and is one major factor to bring improvements in accuracy, general robustness and overconfidence performance. 

Moving forward, it would be interesting to carry out a broader empirical study with other base GNN architectures such as GAT \citep{Velickovic2017-qm} and PNA \citep{Corso2020-tj}.  Another interesting direction, which is already initiated by the ablation study detailed in Appendix \ref{ablation_results_section}, is to find a quantitative way to measure distance-preservation within learned representation and understand the trade-off between the preservation and task-specific compression via the lens of accuracy and robustness performance.

%\balaji{ensure that bib is consistent. check for capitalization of names e.g {B}ayesian vs bayesian. some missing bib info for articles, cite conference versions than arxiv when possible }

\bibliographystyle{plainnat}
\bibliography{./references}

\begin{thebibliography}{32}
\providecommand{\natexlab}[1]{#1}
\providecommand{\url}[1]{\texttt{#1}}
\expandafter\ifx\csname urlstyle\endcsname\relax
  \providecommand{\doi}[1]{doi: #1}\else
  \providecommand{\doi}{doi: \begingroup \urlstyle{rm}\Url}\fi

\bibitem[Bajusz et~al.(2015)Bajusz, R{\'a}cz, and H{\'e}berger]{Bajusz2015-oz}
D{\'a}vid Bajusz, Anita R{\'a}cz, and K{\'a}roly H{\'e}berger.
\newblock Why is {T}animoto index an appropriate choice for fingerprint-based
  similarity calculations?
\newblock \emph{J. Cheminform.}, 7:\penalty0 20, May 2015.

\bibitem[Battaglia et~al.(2016)Battaglia, Pascanu, Lai, Rezende, and
  Kavukcuoglu]{Battaglia2016-qj}
Peter~W Battaglia, Razvan Pascanu, Matthew Lai, Danilo Rezende, and Koray
  Kavukcuoglu.
\newblock {Interaction Networks for Learning about Objects, Relations and
  Physics}.
\newblock In \emph{NeurIPS}, 2016.

\bibitem[{Center for Drug Evaluation} and
  {Research}(2005)]{Center_for_Drug_Evaluation_undated-ql}
{Center for Drug Evaluation} and {Research}.
\newblock {S7B} {Nonclinical Evaluation of the Potential for Delayed
  Ventricular Repolarization (QT Interval Prolongation) by Human
  Pharmaceuticals}, 2005.

\bibitem[Corso et~al.(2020)Corso, Cavalleri, Beaini, Li{\`o}, and Veli{\v
  c}kovi{\'c}]{Corso2020-tj}
Gabriele Corso, Luca Cavalleri, Dominique Beaini, Pietro Li{\`o}, and Petar
  Veli{\v c}kovi{\'c}.
\newblock {Principal Neighbourhood Aggregation for Graph Nets}.
\newblock April 2020.

\bibitem[Duvenaud et~al.(2015)Duvenaud, Maclaurin, Aguilera-Iparraguirre,
  G{\'o}mez-Bombarelli, Hirzel, Aspuru-Guzik, and Adams]{Duvenaud2015-zr}
David Duvenaud, Dougal Maclaurin, Jorge Aguilera-Iparraguirre, Rafael
  G{\'o}mez-Bombarelli, Timothy Hirzel, Al{\'a}n Aspuru-Guzik, and Ryan~P
  Adams.
\newblock {Convolutional Networks on Graphs for Learning Molecular
  Fingerprints}.
\newblock \emph{NeurIPS}, 2015.

\bibitem[Feng et~al.(2020)Feng, Wang, Wang, and Ding]{feng2020uncertainty}
Boyuan Feng, Yuke Wang, Zheng Wang, and Yufei Ding.
\newblock Uncertainty-aware attention graph neural network for defending
  adversarial attacks.
\newblock \emph{arXiv preprint arXiv:2009.10235}, 2020.

\bibitem[Fout et~al.(2017)Fout, Byrd, Shariat, and Ben-Hur]{Fout_undated-yf}
Alex Fout, Jonathon Byrd, Basir Shariat, and Asa Ben-Hur.
\newblock Protein interface prediction using graph convolutional networks.
\newblock \emph{NeurIPS}, 2017.

\bibitem[Gaulton et~al.(2012)Gaulton, Bellis, Bento, Chambers, Davies, Hersey,
  Light, McGlinchey, Michalovich, Al-Lazikani, and Overington]{Gaulton2012-og}
Anna Gaulton, Louisa~J Bellis, A~Patricia Bento, Jon Chambers, Mark Davies,
  Anne Hersey, Yvonne Light, Shaun McGlinchey, David Michalovich, Bissan
  Al-Lazikani, and John~P Overington.
\newblock {ChEMBL}: a large-scale bioactivity database for drug discovery.
\newblock \emph{Nucleic Acids Res.}, 40\penalty0 (Database issue):\penalty0
  D1100--7, January 2012.

\bibitem[Geisler et~al.(2020)Geisler, Z{\"u}gner, and
  G{\"u}nnemann]{geisler2020reliable}
Simon Geisler, Daniel Z{\"u}gner, and Stephan G{\"u}nnemann.
\newblock Reliable graph neural networks via robust aggregation.
\newblock \emph{arXiv preprint arXiv:2010.15651}, 2020.

\bibitem[Gilmer et~al.(2017)Gilmer, Schoenholz, Riley, Vinyals, and
  Dahl]{Gilmer2017-yn}
Justin Gilmer, Samuel~S Schoenholz, Patrick~F Riley, Oriol Vinyals, and
  George~E Dahl.
\newblock {Neural Message Passing for Quantum Chemistry}.
\newblock \emph{ICML}, 2017.

\bibitem[Gouk et~al.(2021)Gouk, Frank, Pfahringer, and Cree]{Gouk2018-ia}
Henry Gouk, Eibe Frank, Bernhard Pfahringer, and Michael~J Cree.
\newblock Regularisation of neural networks by enforcing {L}ipschitz
  continuity.
\newblock \emph{Machine Learning}, 110\penalty0 (2):\penalty0 393--416, 2021.

\bibitem[Gulrajani et~al.(2017)Gulrajani, Ahmed, Arjovsky, Dumoulin, and
  Courville]{Gulrajani2017-mu}
Ishaan Gulrajani, Faruk Ahmed, Martin Arjovsky, Vincent Dumoulin, and Aaron
  Courville.
\newblock Improved training of {W}asserstein {GANs}.
\newblock \emph{NeurIPS}, 2017.

\bibitem[Hirschfeld et~al.(2020)Hirschfeld, Swanson, Yang, Barzilay, and
  Coley]{Hirschfeld2020-lt}
Lior Hirschfeld, Kyle Swanson, Kevin Yang, Regina Barzilay, and Connor~W Coley.
\newblock {Uncertainty Quantification Using Neural Networks for Molecular
  Property Prediction}.
\newblock \emph{J. Chem. Inf. Model.}, 60\penalty0 (8):\penalty0 3770--3780,
  August 2020.

\bibitem[Hwang et~al.(2020)Hwang, Lee, Jo, Yoon, and Ryu]{Hwang2020-mz}
Doyeong Hwang, Grace Lee, Hanseok Jo, Seyoul Yoon, and Seongok Ryu.
\newblock A benchmark study on reliable molecular supervised learning via
  {B}ayesian learning.
\newblock \emph{arXiv preprint arXiv:2006.07021}, 2020.

\bibitem[Jacobsen et~al.(2018)Jacobsen, Smeulders, and
  Oyallon]{Jacobsen2018-sj}
J{\"o}rn-Henrik Jacobsen, Arnold Smeulders, and Edouard Oyallon.
\newblock {i-RevNet}: {Deep Invertible Networks}.
\newblock \emph{ICLR}, 2018.

\bibitem[Kearnes et~al.(2016)Kearnes, McCloskey, Berndl, Pande, and
  Riley]{Kearnes2016-sk}
Steven Kearnes, Kevin McCloskey, Marc Berndl, Vijay Pande, and Patrick Riley.
\newblock Molecular graph convolutions: moving beyond fingerprints.
\newblock \emph{J. Comput. Aided Mol. Des.}, 30\penalty0 (8):\penalty0
  595--608, August 2016.

\bibitem[Lakshminarayanan et~al.(2017)Lakshminarayanan, Pritzel, and
  Blundell]{lakshminarayanan2016simple}
Balaji Lakshminarayanan, Alexander Pritzel, and Charles Blundell.
\newblock Simple and scalable predictive uncertainty estimation using deep
  ensembles.
\newblock \emph{NeurIPS}, 2017.

\bibitem[Liu et~al.(2020)Liu, Lin, Padhy, Tran, Bedrax-Weiss, and
  Lakshminarayanan]{liu2020simple}
Jeremiah~Zhe Liu, Zi~Lin, Shreyas Padhy, Dustin Tran, Tania Bedrax-Weiss, and
  Balaji Lakshminarayanan.
\newblock Simple and principled uncertainty estimation with deterministic deep
  learning via distance awareness.
\newblock In \emph{NeurIPS}, 2020.

\bibitem[Lu et~al.(2020)Lu, Ie, and Sha]{lu2020uncertainty}
Zhiyun Lu, Eugene Ie, and Fei Sha.
\newblock Uncertainty estimation with infinitesimal jackknife, its distribution
  and mean-field approximation.
\newblock \emph{arXiv preprint arXiv:2006.07584}, 2020.

\bibitem[McCloskey et~al.(2020)McCloskey, Sigel, Kearnes, Xue, Tian, Moccia,
  Gikunju, Bazzaz, Chan, Clark, Cuozzo, Gui{\'e}, Guilinger, Huguet, Hupp,
  Keefe, Mulhern, Zhang, and Riley]{McCloskey2020-es}
Kevin McCloskey, Eric~A Sigel, Steven Kearnes, Ling Xue, Xia Tian, Dennis
  Moccia, Diana Gikunju, Sana Bazzaz, Betty Chan, Matthew~A Clark, John~W
  Cuozzo, Marie-Aude Gui{\'e}, John~P Guilinger, Christelle Huguet,
  Christopher~D Hupp, Anthony~D Keefe, Christopher~J Mulhern, Ying Zhang, and
  Patrick Riley.
\newblock Machine {L}earning on {DNA-Encoded} {L}ibraries: {A New Paradigm for
  Hit Finding}.
\newblock \emph{J. Med. Chem.}, June 2020.

\bibitem[Miyato et~al.(2018)Miyato, Kataoka, Koyama, and
  Yoshida]{Miyato2018-bg}
Takeru Miyato, Toshiki Kataoka, Masanori Koyama, and Yuichi Yoshida.
\newblock Spectral normalization for generative adversarial networks.
\newblock In \emph{ICLR}, 2018.

\bibitem[Ovadia et~al.(2019)Ovadia, Fertig, Ren, Nado, Sculley, Nowozin,
  Dillon, Lakshminarayanan, and Snoek]{Ovadia2019-by}
Yaniv Ovadia, Emily Fertig, Jie Ren, Zachary Nado, D~Sculley, Sebastian
  Nowozin, Joshua~V Dillon, Balaji Lakshminarayanan, and Jasper Snoek.
\newblock {Can You Trust Your Model's Uncertainty? Evaluating Predictive
  Uncertainty Under Dataset Shift}.
\newblock \emph{NeurIPS}, 2019.

\bibitem[Rahimi and Recht(2007)]{Rahimi_undated-sq}
Ali Rahimi and Ben Recht.
\newblock Random features for large-scale kernel machines.
\newblock \emph{NeurIPS}, 2007.

\bibitem[Rogers and Hahn(2010)]{Rogers2010-yh}
David Rogers and Mathew Hahn.
\newblock Extended-connectivity fingerprints.
\newblock \emph{J. Chem. Inf. Model.}, 50\penalty0 (5):\penalty0 742--754, May
  2010.

\bibitem[Sanchez-Gonzalez et~al.(2018)Sanchez-Gonzalez, Heess, Springenberg,
  Merel, Riedmiller, Hadsell, and Battaglia]{Sanchez-Gonzalez2018-fs}
Alvaro Sanchez-Gonzalez, Nicolas Heess, Jost~Tobias Springenberg, Josh Merel,
  Martin Riedmiller, Raia Hadsell, and Peter Battaglia.
\newblock Graph networks as learnable physics engines for inference and
  control.
\newblock In \emph{ICML}, 2018.

\bibitem[Siramshetty et~al.(2020)Siramshetty, Nguyen, Martinez, Southall,
  Simeonov, and Zakharov]{Siramshetty2020-yy}
Vishal~B Siramshetty, Dac-Trung Nguyen, Natalia~J Martinez, Noel~T Southall,
  Anton Simeonov, and Alexey~V Zakharov.
\newblock {Critical Assessment of Artificial Intelligence Methods for
  Prediction of {hERG} Channel Inhibition in the ``Big Data'' Era}.
\newblock \emph{J. Chem. Inf. Model.}, December 2020.

\bibitem[Smith et~al.(1996)Smith, Baukrowitz, and Yellen]{Smith1996-df}
P~L Smith, T~Baukrowitz, and G~Yellen.
\newblock The inward rectification mechanism of the {HERG} cardiac potassium
  channel.
\newblock \emph{Nature}, 379\penalty0 (6568):\penalty0 833--836, February 1996.

\bibitem[Van~Amersfoort et~al.(2020)Van~Amersfoort, Smith, Teh, and
  Gal]{Van_Amersfoort2020-gg}
Joost Van~Amersfoort, Lewis Smith, Yee~Whye Teh, and Yarin Gal.
\newblock Uncertainty estimation using a single deep deterministic neural
  network.
\newblock In \emph{ICML}, 2020.

\bibitem[van Amersfoort et~al.(2021)van Amersfoort, Smith, Jesson, Key, and
  Gal]{Van_Amersfoort2021-qt}
Joost van Amersfoort, Lewis Smith, Andrew Jesson, Oscar Key, and Yarin Gal.
\newblock On feature collapse and deep kernel learning for single forward pass
  uncertainty.
\newblock \emph{arXiv preprint arXiv:2102.11409}, 2021.

\bibitem[Vandenberg et~al.(2012)Vandenberg, Perry, Perrin, Mann, Ke, and
  Hill]{Vandenberg2012-gr}
Jamie~I Vandenberg, Matthew~D Perry, Mark~J Perrin, Stefan~A Mann, Ying Ke, and
  Adam~P Hill.
\newblock {hERG} k(+) channels: structure, function, and clinical significance.
\newblock \emph{Physiol. Rev.}, 92\penalty0 (3):\penalty0 1393--1478, July
  2012.

\bibitem[Veli{\v c}kovi{\'c} et~al.(2017)Veli{\v c}kovi{\'c}, Cucurull,
  Casanova, Romero, Li{\`o}, and Bengio]{Velickovic2017-qm}
Petar Veli{\v c}kovi{\'c}, Guillem Cucurull, Arantxa Casanova, Adriana Romero,
  Pietro Li{\`o}, and Yoshua Bengio.
\newblock {Graph Attention Networks}.
\newblock October 2017.

\bibitem[Wu et~al.(2018)Wu, Ramsundar, Feinberg, Gomes, Geniesse, Pappu,
  Leswing, and Pande]{Wu2017-yv}
Zhenqin Wu, Bharath Ramsundar, Evan~N Feinberg, Joseph Gomes, Caleb Geniesse,
  Aneesh~S Pappu, Karl Leswing, and Vijay Pande.
\newblock {MoleculeNet}: a benchmark for molecular machine learning.
\newblock \emph{Chemical science}, 9\penalty0 (2):\penalty0 513--530, 2018.

\end{thebibliography}

\clearpage
\newpage

\appendix
\setcounter{figure}{0}
\setcounter{table}{0}
\makeatletter 
\renewcommand{\thefigure}{S\@arabic\c@figure}
\renewcommand{\thetable}{S\@arabic\c@table}
\makeatother

\section{GNN-SNGP architecture}
\label{architecture}

\begin{figure}[ht]
    \centering
    \includegraphics[width=0.8\linewidth]{"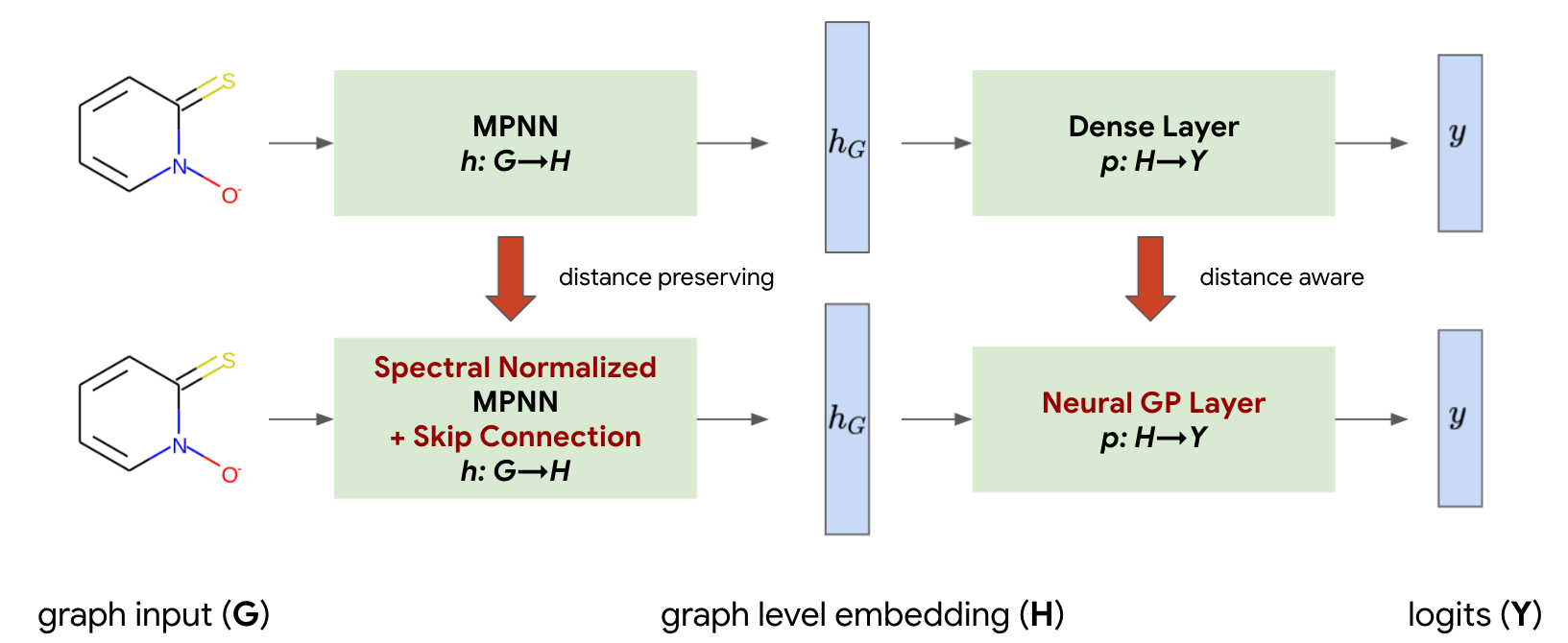"}
    \caption{Improving GNN architecture via distance-preserving feature extractor (skip connection and spectral normalization) and distance-aware classifier (neural Gaussian Processes layer).}
    \label{fig:gnn-gp-architecture1}
\end{figure}

We present high-level architectural changes based on GNN baseline model in Figure \ref{fig:gnn-gp-architecture1}: distance-preserving feature extractor by adding skip connection and spectral normalization regularization, distance aware classifier using neural GP layer.

\section{Evaluation Metrics}
\label{metric_def}

In this study, we examine for each  accuracy performance, general robustness performance, overconfidence performance as well as distance-awareness performance.

\paragraph{Accuracy performance} AUROC for the classification benchmark.

\paragraph{Robustness performance} Expected Calibration Error (abbr. ECE), Brier Score (abbr. Brier), and Negative Log Likelihood (abbr. NLL).

\paragraph{Overconfidence performance: OFNs\% and OFPs\%} Occurance of overconfident false negatives and false positives. Depending on actual applications, we may care more about Overconfident False Negatives (OFNs) in the drug cardiotoxicity task whereas in the molHIV task we care about general overconfidence, therefore both OFNs and OFPs (Overconfident False Positives). In this study, an overconfident misprediction is defined as any test sample whose predictive confidence (computed via Maximum Softmax Probability, i.e., $\max(p_k)$ is higher than 90\% yet its prediction is wrong, namely:

\begin{equation}
    \max_k(p_k) > 90\% \hspace{0.5cm} \mathsf{and} \hspace{0.5cm}  \arg\max_k (p_k) \neq y^\mathsf{true}
\end{equation}

where $k \in \{0, 1\}$ is class logit index, and $y^\mathsf{true} \in \{0, 1\}$ is the ground truth class.

In this study we measure OFNs\% and OFPs\%, defined as percentages as follows: 
\begin{equation}
    \textnormal{OFNs}\% = \frac{\sum_i \mathbbm{1}(\hat{y}_i<0.1 \hspace{0.1cm} \mathsf{and} \hspace{0.1cm} y_i^\mathsf{true}=1)}{\sum_i \mathbbm{1}(\hat{y}_i<0.1)}
\end{equation}
\begin{equation}
    \textnormal{OFPs}\% = \frac{\sum_i \mathbbm{1}(\hat{y}_i>0.9 \hspace{0.1cm} \mathsf{and} \hspace{0.1cm} y_i^\mathsf{true}=0)}{\sum_i \mathbbm{1}(\hat{y}_i>0.9)}
\end{equation}
where $\hat{y}_i$ is $i$-th sample's predictive probability for positive class, i.e., $p_{k=1}^i$.

\paragraph{Distance-awareness performance: DA-AUC} We design a classification task for a given test set: any molecule belonging to the \textit{close set} (short distance to Train set, defined in \ref{cardiotox_split}) gets ground-truth label 0, any molecule belonging to the \textit{far set} (long distance to Train set, defined in \ref{cardiotox_split}) gets ground truth label 1. We use predictive uncertainty (computed via 1-max$(p_k)$) to classify if a test sample is in the \textit{close set} or the \textit{far set}. DA-AUC is the AUROC measurement for this task.

\section{Distance analysis for OFNs in CardioTox}
\begin{figure}[h]
\centering
    \begin{subfigure}{.55\textwidth}
      \centering
      \includegraphics[width=\linewidth]{"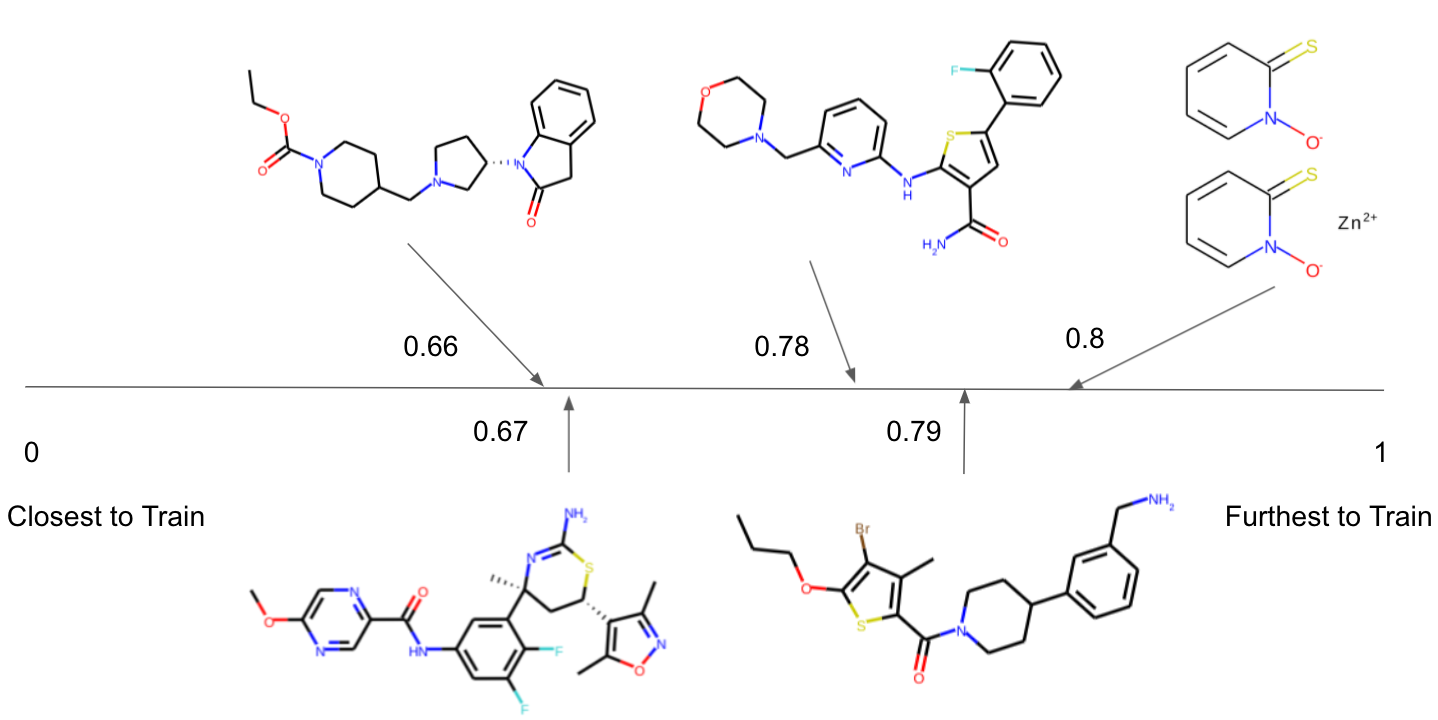"}
      \caption{}
      \label{fig:example_ofns}
    \end{subfigure}%
    \begin{subfigure}{.4\textwidth}
      \centering
      \includegraphics[width=\linewidth]{"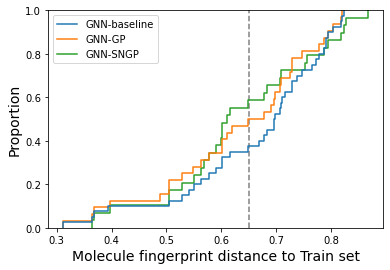"}
      \caption{}
      \label{fig:ofn_distance_cdf}
    \end{subfigure}
    \vspace{-0.5em}
    \caption{Graph distance analysis for the overconfident false negatives (OFNs) in CardioTox benchmark. (a) shows five overconfident false negatives with their distance (in 0-1 scale) to the Train set. (b) Cumulative distributions over fingerprint distance from the OFNs (incurred in corresponding models) to the Train set.
}
\label{fig:distant_ofns}
\end{figure}

Here we present our analysis on OFNs incurred in GNN baseline model. One of the outstanding observations is that a good portion of OFNs molecules are distant from the train set. Figure \ref{fig:example_ofns} shows a few such example molecules. More quantitatively, Figure \ref{fig:ofn_distance_cdf} suggests over 60\% of OFNs have Tanimoto distance $>0.65$, which is often regarded as a condition of having novel molecule structure. As we introduce models with more distance-awareness (e.g., GNN-GP, GNN-SNGP), distant molecules becomes less dominant in OFNs.

\section{Drug cardiotoxicity data split}
\label{cardiotox_split}

\begin{figure}[ht]
    \centering
    \includegraphics[width=0.8\linewidth]{"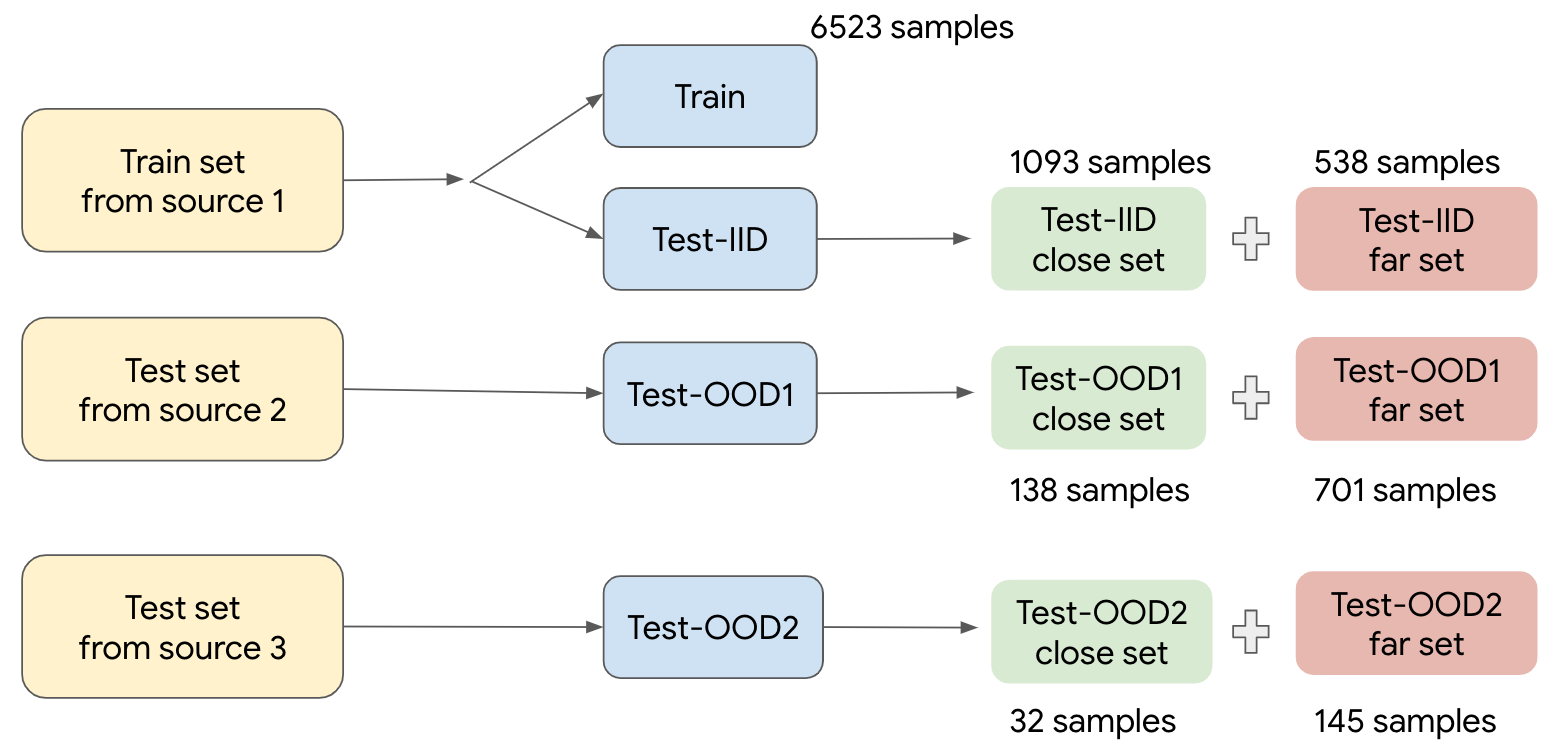"}
    \caption{CardioTox benchmark designed for investigation on accuracy and robustness performance under various levels of distributional shift. Majority of Test-IID are similar molecules to Train set, while 84\% of Test-OOD1 and 82\% of Test-OOD2 have novel structures.}
    \label{fig:data-split}
\end{figure}

We re-organized the original hERG (cardiotoxicity anti-target) dataset \citep{Siramshetty2020-yy} into \textbf{CardioTox} to facilitate graph model reliability research. The original dataset comes with a Train set, a compilation from ChEMBL \citep{Gaulton2012-og} and NCATS (National Center for Advancing Translational Sciences), a prospective validation set from NCATS \citet{Siramshetty2020-yy} and a test set from FDA (see Figure \ref{fig:data-split}). 

As a first step, we added graph structural information such as node features, edge features and adjacency matrix for each molecule record and created three test sets: Test-IID (randomly sampled 20\% from the Train distribution), Test-OOD1 (the prospective validation set from NCATS) and Test-OOD2 (from FDA). The shift is mainly from input distribution (i.e., graph structures): majority of Test-IID are similar molecules to Train set, while 84\% of Test-OOD1 and 82\% of Test-OOD2 have novel structures.

As a second step, we further split each test set into \textit{close set} and \textit{far set}. Using Tanimoto graph distance \citep{Bajusz2015-oz} defined by molecule fingerprint representation \citep{Rogers2010-yh}, we group test samples with distance (average distance to top 8 nearest training samples) < 0.7 into \textit{close set} and remaining samples go to \textit{far set}. Figure \ref{fig:data-split} shows the detailed splitting flow and final counts. 

\section{Related work}
\label{related_work}
On GNN robustness research, \citet{geisler2020reliable} proposed a novel GNN aggregation function, Soft Medoid, to improve robustness against adversarial attacks from edge perturbation. \citet{feng2020uncertainty} constructed a two-branched GNN architecture where the first branch computes model uncertainty and data uncertainty, the second utilizes those uncertainties to adjust attention during aggregation. Those studies are organized around defending adversarial attacks, while our work aims to mitigate overconfidence issue by introducing targeted techniques.

In the area of molecule graph learning, \citet{Hwang2020-mz} have recently applied Deep Ensemble, Monte Carlo dropout, Stochastic Gradient Langevin Dynamics (SGLD), Stochastic Weight Averaging (SWA), and Stochastic Weight Averaging Gaussian (SWAG) to GNN models and those techniques can be categorized as different approaches to create ensemble models. For example, SWA picks ensemble members along the training procedure. Our work focuses on a single GNN model and introduces architectural changes to improve its distance-awareness to reduce overconfident mispredictions. Another related work is from \citet{Hirschfeld2020-lt} where a GNN model is first trained and its latent representation gets fed to a downstream Gaussian Processes. In contrast to that, we develop an end-to-end trainable GNN-SNGP without computational constraint of kernel computation and storage that come with exact Gaussian processes, making GNN-SNGP better suited to processing large scale datasets such as high throughput screening data in drug discovery \citep{McCloskey2020-es}.

On the distance-preservation/awareness side, compared with SNGP \citep{liu2020simple} techniques used in this work, two-sided gradient penalty \citep{Gulrajani2017-mu,Van_Amersfoort2020-gg} constrains gradient deviation by placing penalty on the loss and can be sensitive to tune. Deep Invertible Networks \citep{Jacobsen2018-sj} achieves high level of distance-preservation via invertiable layers but are often difficult to train and expensive in memory.

\section{Uncertainty improvements for OFNs}
\label{ofn_uncertainty_improvement}
\begin{figure}[ht]
    \centering
    \includegraphics[width=0.7\linewidth]{"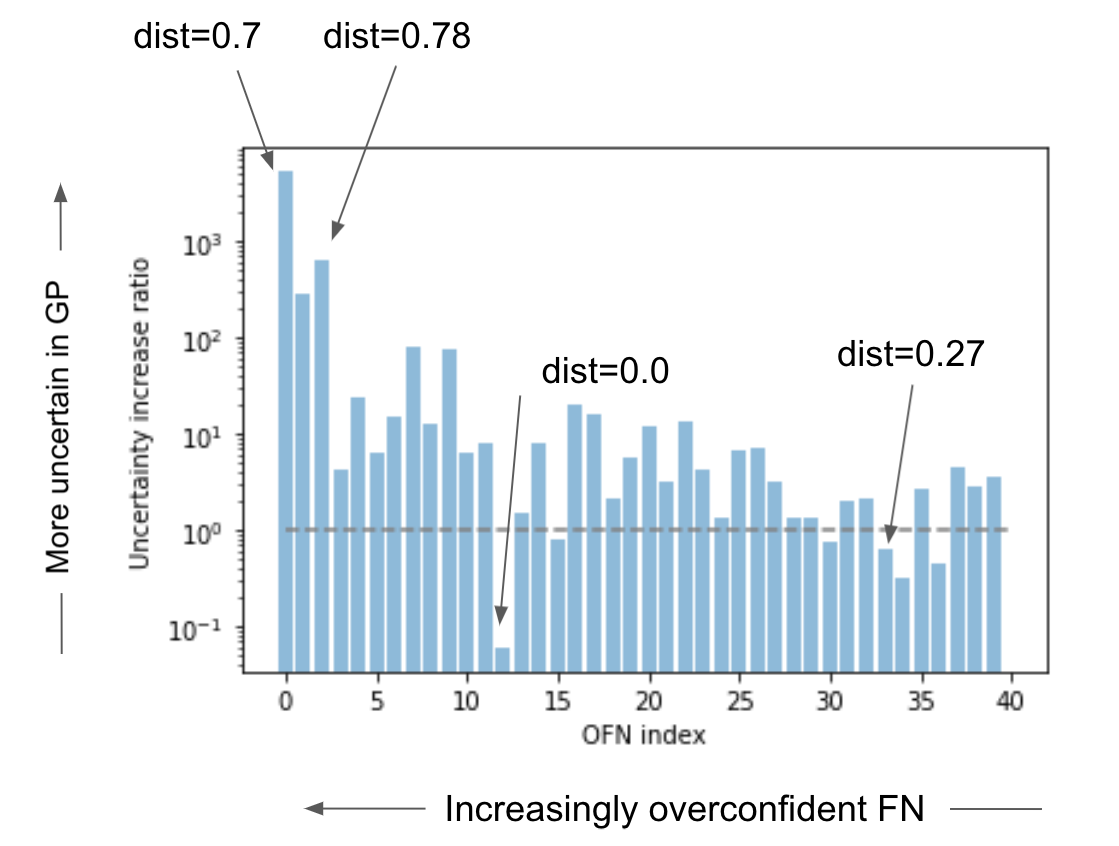"}
    \caption{Uncertainty increase ratio for the top 40 overconfident false negatives. Two distant (distance = 0.7 and 0.78) and two close (distance = 0.0 and 0.27) OFNs are annotated.}
    \label{fig:uncertainty_increase_ratio}
\end{figure}

We have carefully looked into the OFNs generated by the GNN baseline, and assess their uncertainty improvements by monitoring uncertainty increase ratio (UIR):
\begin{equation}
    \textnormal{UIR} = \frac{U_{\textnormal{GNN-GP}}}{U_{\textnormal{GNN baseline}}}    
\end{equation}

where $U_{\textnormal{GNN-GP}}, U_{\textnormal{GNN baseline}}$ are the uncertainty estimates by the GNN-GP model and GNN baseline model. UIR > 1 indicates that the GNN-GP becomes less overconfident than the GNN baseline. Figure \ref{fig:uncertainty_increase_ratio} lists UIRs for the 40 top overconfident false negatives from the GNN baseline. We observe over 80\% of them have improved uncertainty estimates (UIR>1). As annotated in Figure \ref{fig:uncertainty_increase_ratio}, many highly improved OFNs are the ones distant from the Train set, while the ones getting worse uncertainty estimates often are close to the Train set. This is expected in a model with high distance-awareness.

\section{Results on existing benchmarks}
\label{established_benchmarks}
In addition to \textbf{CardioTox}, we have applied our GNN-GP and GNN-SNGP models to three established molecule benchmarks: molHIV, BBBP and BACE. Specifically, molHIV is a dataset of HIV antiviral activity (each molecule has an active or inactive label), BBBP a dataset of Brain-Blood Barrier Penetration (each molecule has a label indicating whether it can penetrate through brain cell membrane to enter central nervous system) and BACE a dataset of binding affinity against human beta-secretase 1 (each molecule has a label indicating whether it binds to human beta-secretase 1). Table \ref{benchmarks_results} shows the results.

\begin{table}[h]
  \caption{Accuracy (AUROC) and robustness (ECE, Brier, NLL) performance for molHIV, BBBP, BACE benchmark. Results are averaged over 10 seeds.
  }
  \label{benchmarks_results}
  \centering
  \resizebox{\textwidth}{!}{  
  \begin{tabular}{lllllll}
    \toprule
    molHIV & AUROC ($\uparrow$) & ECE ($\downarrow$) & Brier ($\downarrow$) & NLL ($\downarrow$) & OFNs\% ($\downarrow$) & OFPs\% ($\downarrow$)  \\
    \midrule
    GNN baseline & 0.738$\pm$0.004 & 0.019$\pm$0.001 & 0.058$\pm$0.002  & 0.143$\pm$0.009 & 2.38$\pm$0.05  & 1.58$\pm$0.40    \\
    GNN-GP     & \textbf{0.756$\pm$0.004} & 0.013$\pm$0.001 & \textbf{0.054$\pm$0.003} & 0.125$\pm$0.006 & \textbf{2.14$\pm$0.05} & 0.16$\pm$0.15      \\
    GNN-SNGP     & 0.745$\pm$0.003 & \textbf{0.008$\pm$0.001} & \textbf{0.055$\pm$0.002} & \textbf{0.122$\pm$0.003} & \textbf{2.16$\pm$0.05} & \textbf{0.00$\pm$0.00}  \\
    \bottomrule
  \end{tabular}
  }
  \resizebox{\textwidth}{!}{  
  \begin{tabular}{lllllll}
    \toprule
    BBBP & AUROC ($\uparrow$) & ECE ($\downarrow$) & Brier ($\downarrow$) & NLL ($\downarrow$) & OFNs\% ($\downarrow$) & OFPs\% ($\downarrow$)  \\
    \midrule
    GNN baseline & 0.773$\pm$0.002  & 0.202$\pm$0.007 & 0.495$\pm$0.029  & 0.911$\pm$0.059 & 29.50$\pm$1.32  & 11.43$\pm$0.88   \\
    GNN-GP     & \textbf{0.792$\pm$0.005} & 0.136$\pm$0.012 & \textbf{0.427$\pm$0.035} & 0.679$\pm$0.066 & \textbf{27.23$\pm$1.44} & 6.05$\pm$0.93      \\
    GNN-SNGP     & 0.776$\pm$0.007 & \textbf{0.119$\pm$0.010} & \textbf{0.426$\pm$0.022} & \textbf{0.651$\pm$0.032} & \textbf{26.18$\pm$4.29} & \textbf{4.33$\pm$0.78}  \\
    \bottomrule
  \end{tabular}
  }
  \resizebox{\textwidth}{!}{  
  \begin{tabular}{lllllll}
    \toprule
    BACE & AUROC ($\uparrow$) & ECE ($\downarrow$) & Brier ($\downarrow$) & NLL ($\downarrow$) & OFNs\% ($\downarrow$) & OFPs\% ($\downarrow$)  \\
    \midrule
    GNN baseline & 0.789$\pm$0.004  & 0.181$\pm$0.011 & 0.460$\pm$0.064  & 0.784$\pm$0.101 & 27.90$\pm$1.71  & 5.77$\pm$0.61    \\
    GNN-GP     & \textbf{0.807$\pm$0.004} & 0.145$\pm$0.010 & \textbf{0.426$\pm$0.034} & 0.674$\pm$0.074 & 24.15$\pm$1.79 & 2.24$\pm$0.46    \\
    GNN-SNGP     &0.803$\pm$0.003 & \textbf{0.105$\pm$0.013} & \textbf{0.435$\pm$0.051} & \textbf{0.642$\pm$0.073} & \textbf{21.03$\pm$2.08} & \textbf{0.30$\pm$0.19}  \\
    \bottomrule
  \end{tabular}
  }
\end{table}

% \paragraph{Model hyperparameters and compute resources}
% For all the tasks performed in this study, we use GPU P100 as the compute resource. Each experiment we run 10 replicas to get performance statistics. GNN baseline uses same set of hyperparameters across tasks: message hidden dimension = 32, graph level embedding dimension = 32, learning rate = 0.001, batch size = 128. GP-layer specific hyperparameters: random feature vector dimension = 100, mean field factor uses 0.1 except BBBP where mean field factor = 5.0.

\section{Ablation study on performance contribution}
\label{ablation_results_section}
Distance-preservation in representation learning and distance-awareness in classifier both impact the performances we care about in this study. In order to understand relative contributions of them to the performance improvement, we further carried out an extensive ablation study. We take the latent representations learned by GNN baseline, GNN-GP and GNN-SNGP (increasing distance-preservation) and feed them to classifiers with increasing levels of distance-awareness: Dense layer (using logistic regression), GP-layer, exact Gaussian processes classifier (GPC hereafter). We also added experiments using pre-defined representation (i.e., molecule fingerprint denoted as FPs) which should provide 100\% of distance preservation. Table \ref{ablation_results} shows the results. 

We find a GNN model achieves best performances when both distance-preservation (via SN) and distance-awareness (via GP) are present in the architecture (i.e., GNN-SNGP embeddings with GPC). However, one interesting note is that naively increasing distance-preservation does not guarantee generalization, either in-domain or OOD. For example, we observe that the pre-defined representation (i.e., molecule fingerprint denoted as FPs) offers perfect distance-preservation, but tends to give relatively low AUROC (compared with neural representation) under data shifts such as Test-OOD2 even with exact GPC as its classifier. We hypothesis that this is related to a general trade-off in representation learning between dimension reduction and information preservation. The dimension reduction aspect tries to discard information noisy and/or less relevant to prediction task so that given limited data, models could escape the curse of dimensionality and achieve good performance under finite data. However, this could also lead to the feature collapse phenomenon that harms model robustness especially under distributional shifts. On the other hand, the information preservation aspect recovers robustness by mitigating feature collapse, but keeping around noisy/irrelevant features may demand more data for convergence and thus negatively impact accuracy given limited data. This work tries to find a good balance between these two aspects by designing a distance-preserving representation that can be optimized toward the task at hand. 
% In addition, neural representations with similar degree distance-preservation may still display different levels of suitability to the prediction task. Finding representations suitable for the prediction task among those with high distance-preservation is another key goal of this study.

\begin{table}[h]
  \caption{Ablation study results for Accuracy (AUROC), robustness (ECE, Brier, NLL) and overconfidence (OFNs) performance for CardioTox benchmark. Note that since Dense (implemented by logistic regression) and GPC use deterministic optimization instead of stochastic gradient descent, the error bars are zero.}
  \label{ablation_results}
  \centering
  \resizebox{\textwidth}{!}{ 
  \begin{tabular}{lllllll}
    \toprule
    Test-IID & AUROC ($\uparrow$) & ECE ($\downarrow$) & Brier ($\downarrow$) & NLL ($\downarrow$) & OFNs\% ($\downarrow$) \\
    \midrule
    % FPs + GPC & \textbf{0.921}  & 0.074 & 0.216  & 0.355 & \textbf{2.07} \\
    FPs + GPC & \textbf{0.921$\pm$0.000}  & 0.024$\pm$0.000 & \textbf{0.201$\pm$0.000}  & \textbf{0.332$\pm$0.000} & 4.74$\pm$0.00 \\
    FPs + GP-layer & 0.881$\pm$0.001  & \textbf{0.020$\pm$0.003} & 0.250$\pm$0.012  & 0.396$\pm$0.015 & 4.97$\pm$0.46 \\
    FPs + Dense & 0.895$\pm$0.000  & 0.065$\pm$0.000 & 0.221$\pm$0.000  & 0.436$\pm$0.000 & 6.43$\pm$0.00 \\
    \midrule
    GNN-baseline embed + GPC     & 0.923$\pm$0.000 & 0.035$\pm$0.000 & 0.181$\pm$0.000 & 0.335$\pm$0.000 & 4.29$\pm$0.00     \\
    GNN-baseline embed + GP-layer     & 0.920$\pm$0.001 & 0.037$\pm$0.002 & 0.189$\pm$0.003 & 0.328$\pm$0.005 & 4.36$\pm$0.00      \\
    GNN-baseline embed + Dense & 0.919$\pm$0.000  & 0.037$\pm$0.000 & 0.194$\pm$0.000  & 0.352$\pm$0.000 & 4.05$\pm$0.00 \\
    GNN-GP embed + GPC     & 0.934$\pm$0.000 & 0.005$\pm$0.000 & \textbf{0.169$\pm$0.000} & 0.282$\pm$0.000 & 2.98$\pm$0.00 \\
    GNN-GP embed + GP-layer     & 0.933$\pm$0.001 & 0.033$\pm$0.002 & 0.177$\pm$0.003 & 0.298$\pm$0.007 & 3.77$\pm$0.00      \\
    GNN-GP embed + Dense     & 0.904$\pm$0.000 & 0.042$\pm$0.000 & 0.217$\pm$0.000 & 0.377$\pm$0.000 & 4.37$\pm$0.00 \\
    GNN-SNGP embed + GPC     & \textbf{0.935$\pm$0.000} & \textbf{0.002$\pm$0.000} & 0.172$\pm$0.000 & \textbf{0.280$\pm$0.000} & \textbf{2.92$\pm$0.00} \\
    GNN-SNGP embed + GP-layer     & 0.932$\pm$0.001 & 0.028$\pm$0.001 & 0.179$\pm$0.005 & 0.295$\pm$0.005 & 3.56$\pm$0.21 \\
    GNN-SNGP embed + Dense     & 0.894$\pm$0.000 & 0.021$\pm$0.000 & 0.223$\pm$0.000 & 0.375$\pm$0.000 & 4.63$\pm$0.00 \\
    \bottomrule
    \toprule
    Test-OOD1 &  AUROC ($\uparrow$) & ECE ($\downarrow$) & Brier ($\downarrow$) & NLL ($\downarrow$) & OFNs\% ($\downarrow$) \\
    \midrule
    % FPs + GPC & \textbf{0.834}  & 0.217 & 0.210  & 0.372 & \textbf{0.00} \\
    FPs + GPC & \textbf{0.834$\pm$0.000}  & 0.097$\pm$0.000 & \textbf{0.133$\pm$0.000}  & \textbf{0.248$\pm$0.000} & 1.21$\pm$0.00 \\
    FPs + GP-layer & 0.764$\pm$0.009  & \textbf{0.076$\pm$0.010} & 0.192$\pm$0.099  & 0.325$\pm$0.124 & 1.87$\pm$0.39 \\
    FPs + Dense & 0.761$\pm$0.000  & 0.112$\pm$0.000 & 0.326$\pm$0.000  & 0.611$\pm$0.000 & 2.67$\pm$0.00 \\
    \midrule
    GNN-baseline embed + GPC      & 0.794$\pm$0.000 & 0.069$\pm$0.000 & 0.290$\pm$0.000 & 0.487$\pm$0.000 & 1.83$\pm$0.00   \\
    GNN-baseline embed + GP-layer     & 0.796$\pm$0.001 & 0.094$\pm$0.006 & 0.334$\pm$0.034 & 0.540$\pm$0.052 & 1.47$\pm$0.05      \\
    GNN-baseline embed + Dense & 0.786$\pm$0.000  & 0.102$\pm$0.000 & 0.343$\pm$0.000  & 0.578$\pm$0.000 & 1.68$\pm$0.00  \\
    GNN-GP embed + GPC      & 0.805$\pm$0.000 & 0.050$\pm$0.000 & 0.279$\pm$0.000 & 0.463$\pm$0.000 & 1.48$\pm$0.00  \\
    GNN-GP embed + GP-layer     & 0.810$\pm$0.001 & 0.072$\pm$0.004 & 0.295$\pm$0.023 & 0.472$\pm$0.039 & 1.55$\pm$0.03      \\
    GNN-GP embed + Dense      & 0.773$\pm$0.000 & 0.084$\pm$0.000 & 0.315$\pm$0.000 & 0.523$\pm$0.000 & 2.15$\pm$0.00 \\
    GNN-SNGP embed + GPC     & \textbf{0.841$\pm$0.000} & \textbf{0.034$\pm$0.000} & \textbf{0.270$\pm$0.000} & \textbf{0.429$\pm$0.000} & 1.17$\pm$0.00 \\
    GNN-SNGP embed + GP-layer    & 0.836$\pm$0.003 & 0.074$\pm$0.008 & 0.316$\pm$0.047 & 0.503$\pm$0.072 & 1.31$\pm$0.09 \\
    GNN-SNGP embed + Dense     & 0.769$\pm$0.000 & 0.073$\pm$0.000 & 0.324$\pm$0.000 & 0.506$\pm$0.000 & \textbf{1.16$\pm$0.00} \\
    \bottomrule
    \toprule
    Test-OOD2 &  AUROC ($\uparrow$) & ECE ($\downarrow$) & Brier ($\downarrow$) & NLL ($\downarrow$) & OFNs\% ($\downarrow$) \\
    \midrule
    % FPs + GPC & \textbf{0.784}  & 0.187 & 0.231  & 0.390 & \textbf{0.00} \\
    FPs + GPC & \textbf{0.784$\pm$0.000}  & 0.074$\pm$0.000 & \textbf{0.167$\pm$0.000}  & \textbf{0.287$\pm$0.000} & \textbf{0.00$\pm$0.00} \\
    FPs + GP-layer & 0.716$\pm$0.016  & \textbf{0.051$\pm$0.004} & 0.219$\pm$0.080  & 0.357$\pm$0.101 & 4.10$\pm$0.55 \\
    FPs + Dense & 0.604$\pm$0.000  & 0.176$\pm$0.000 & 0.397$\pm$0.000  & 0.783$\pm$0.000 & 5.71$\pm$0.00 \\
    \midrule
    GNN-baseline embed + GPC      & 0.842$\pm$0.000 & 0.081$\pm$0.000 & 0.265$\pm$0.000 & 0.451$\pm$0.000 & 1.89$\pm$0.00  \\
    GNN-baseline embed + GP-layer     & 0.838$\pm$0.004 & 0.081$\pm$0.006 & 0.266$\pm$0.028 & 0.426$\pm$0.042 & 1.95$\pm$0.14      \\
    GNN-baseline embed + Dense & 0.831$\pm$0.000  & 0.082$\pm$0.000 & 0.284$\pm$0.000 & 0.492$\pm$0.000 & 1.73$\pm$0.00 \\
    GNN-GP embed + GPC      & 0.853$\pm$0.000 & 0.024$\pm$0.000 & 0.253$\pm$0.000 & 0.434$\pm$0.000 & \textbf{0.00$\pm$0.00} \\
    GNN-GP embed + GP-layer     & 0.850$\pm$0.003 & 0.064$\pm$0.004 & 0.257$\pm$0.018 & 0.417$\pm$0.032 & 1.42$\pm$0.15      \\
    GNN-GP embed + Dense      & 0.825$\pm$0.000 & 0.087$\pm$0.000 & 0.284$\pm$0.000 & 0.512$\pm$0.000 & 2.04$\pm$0.00 \\
    GNN-SNGP embed + GPC     & \textbf{0.891$\pm$0.000} & \textbf{0.021$\pm$0.000} & \textbf{0.208$\pm$0.000} & \textbf{0.347$\pm$0.000} & \textbf{0.00$\pm$0.00} \\
    GNN-SNGP embed + GP-layer     & 0.885$\pm$0.007 & 0.044$\pm$0.006 & 0.238$\pm$0.040 & 0.389$\pm$0.068 & 1.02$\pm$0.11  \\
    GNN-SNGP embed + Dense     & 0.824$\pm$0.000 & 0.050$\pm$0.000 & 0.253$\pm$0.000 & 0.414$\pm$0.000 & 1.10$\pm$0.00 \\
    \bottomrule
  \end{tabular}
  }
\end{table}

\end{document}